\RequirePackage{fix-cm}
\documentclass[smallextended]{svjour3}       
\smartqed  
\usepackage{graphicx}
\usepackage{makecell}
\usepackage{multirow}
\usepackage[dvipsnames]{xcolor}
\usepackage{filecontents}
\usepackage{pgfplots}
\usepackage{caption}

\usepackage{lscape}
\graphicspath{ {./images/} }

\begin{document}

\title{Tag-less Back-Translation}

\dedication{This work is supported by the National Information Technology Development Agency under the National Information Technology Development Fund PhD Scholarship Scheme 2018}

\author{Idris Abdulmumin \and Bashir Shehu Galadanci \and Aliyu Garba}

\institute{Idris Abdulmumin \at
              Computer Science, Ahmadu Bello University, Zaria, Kaduna, Nigeria. \\
              Tel.: +234-806-295-5509\\
              \email{iabdulmumin@abu.edu.ng}
           \and
           Bashir Shehu Galadanci \at
              Software Engineering, Bayero University, Kano, Kano, Nigeria. \\
              \email{bsgaladanci.se@buk.edu.ng}
		\and
		Aliyu Garba \at
              Computer Science, Ahmadu Bello University, Zaria, Kaduna, Nigeria. \\
              \email{algarba@abu.edu.ng}
}

\date{Received: date / Accepted: date}

\maketitle

\begin{abstract}
An effective method to generate a large number of parallel sentences for training improved neural machine translation (NMT) systems is the use of the back-translations of the target-side monolingual data. The standard back-translation method has been shown to be unable to efficiently utilize the available huge amount of existing monolingual data because of the inability of translation models to differentiate between the authentic and synthetic parallel data during training. Tagging, or using gates, has been used to enable translation models to distinguish between synthetic and authentic data, improving standard back-translation and also enabling the use of iterative back-translation on language pairs that underperformed using standard back-translation. In this work, we approach back-translation as a domain adaptation problem, eliminating the need for explicit tagging. In the approach -- \emph{tag-less back-translation} -- the synthetic and authentic parallel data are treated as out-of-domain and in-domain data respectively and, through pre-training and fine-tuning, the translation model is shown to be able to learn more efficiently from them during training. Experimental results have shown that the approach outperforms the standard and tagged back-translation approaches on low resource English-Vietnamese and English-German neural machine translation.

\end{abstract}

\section{Introduction}
\label{intro}
Neural Machine Translation (NMT) \cite{Bahdanau2014,Gehring2017,Vaswani2017} has been the state-of-the-art approach for machine translation in recent years \cite{Edunov2018,Ott2018}, outperforming Phrase-Based Statistical Machine Translation \cite{Koehn2003} when qualitative parallel data between the languages is available in abundance \cite{Zoph2016}. This training dataset is usually scarce and expensive to compile for many language pairs. Recently, researchers have proposed methods to exploit the easier-to-get monolingual data of one or both of the languages to augment the available parallel data and improve the performance of the translation models. Such methods include integrating a language model \cite{Gulcehre2017}, back-translation \cite{Sennrich2016a,Hoang2018,Graca2019}, forward translation \cite{Zhang2016} and dual learning \cite{He:2016:DLM:3157096.3157188}. The back-translation approach is simple and has been the most effective technique yet for NMT \cite{Edunov2018,Hoang2018}. The method involves training a target-to-source (backward) model on the available authentic bitext. The backward model is then used to translate a large amount of monolingual sentences in the target language into synthetic source sentences, generating the synthetic parallel data. The authentic and synthetic parallel data are then mixed to train a source-to-target (forward) model.

It has been shown that as the amount of monolingual data used in back-translation continues to increase, a point is reached when the model stops learning useful representation and, therefore, the performance of the model starts to drop. This is because the usually noise-infested synthetic data starts to overwhelm the authentic data and the model starts to completely unlearn the correct parameters it learns from the authentic training data \cite{Fadaee2018}. Extensive studies by \cite{Edunov2018} have shown that in low resource NMT, noising beam search outputs improve the models more than other generation methods such as sampling. The authors claimed that the method enhances source-side diversity. But the works of \cite{Caswell2019,Yang2019} found that the noising technique is only a form of tagging, indicating to the model that the noised data is back-translated, enabling it to treat the synthetic data as belonging to a different domain. The model then learns different representations, optimally, from the two data. They, instead, introduced the use of explicit tags (and gates) to indicate synthetic inputs. The tagging approach was shown to outperform the standard back-translation.

In this work, we approach back-translation as a domain adaptation problem, simplifying the works of \cite{Edunov2018,Caswell2019,Yang2019} that explicitly differentiate between the two data using noise/tags/gates. Instead of tagging the synthetic data, our approach -- the \emph{tag-less back-translation} -- aims to enable the model to learn efficiently from the two data through pre-training and fine-tuning. Instead of relying on the model to differentiate between the data, we used the synthetic data as generic domain (out-of-domain) and pre-train the model on this data. We then used the authentic data as in-domain to fine-tune the pre-trained translation model. We hypothesize that although the tagging and noising approaches improve the forward models, our domain-adaptation-tailored approach will provide a flexible method of maximizing the gains in the quantity of the synthetic data and efficiently utilizing the quality in the authentic parallel data. The approach will also enable the use of different training settings on the different data, as obtainable in domain adaptation strategies. It also gets better as more research and more improved ways of domain adaption are proposed.

In domain adaptation, the generic model is not always expected to perform very well in the domain it is to be deployed, hence the model is fine-tuned with a usually smaller but in-domain data. In many languages, the in-domain data is usually low-resourced or non-existent: having the same issue as in low resource neural machine translation. The in-domain data in itself is not sufficient to create a good model while the more abundant out-of-domain data performs poorly when deployed in the target domain. Mixing the two data results in the in-domain data to be lost in the out-of-domain data and the resulting model is not able to also perform well in the target domain. The larger out-of-domain data is, therefore, used to pre-train a model and the weights of this model are used to initialize the training of the in-domain translation model -- a technique referred to as \emph{fine-tuning} \cite{Chu2018}. When a different language pair is used for pre-training than that used during fine-tuning, the approach is regarded to as transfer learning \cite{Zoph2016,Kocmi2018}.

We make the following contributions in this paper:
\renewcommand{\labelitemi}{\textbullet}
\begin{itemize}
	\item we proposed a novel approach that enables a translation model that is trained on synthetic and authentic parallel data to be able to efficiently learn from the the two data, utilizing the different advantages presented by each.
	\item we successfully applied pre-training and fine-tuning to enable the forward model in back-translation to differentiate between synthetic and authentic data during training, achieving a superior performance to standard and the successful tagged back-translation approaches,
	\item experimental results have shown that the approach is superior to the standard and tagging back-translation approaches in low resource English-Vietnamese and English-German neural machine translation systems.
\end{itemize}

The remaining sections are as follows: Section \ref{lit} reviews relevant literature on NMT, leveraging monolingual data in NMT and pre-training and fine-tuning. Section \ref{method} explains the tag-less back-translation approach, Section \ref{exp} describes the data and experimental set-up used in training the models, Section \ref{result} discusses the results obtained after the experiment. We discuss further the findings in Section \ref{discuss}. Finally, in Section \ref{conclusion}, we concluded the work and suggest future directions.

\section{Related Works}
\label{lit}
This section presents prior work on NMT, back-translation and pre-training in NMT.

\subsection{Neural Machine Translation (NMT)}
\label{lit-nmt}
The NMT is based on a sequence-to-sequence encoder-decoder system with attention mechanism \cite{Bahdanau2014,Sutskever2014,Luong2015}. The encoders and decoders are made of neural networks that model the conditional probability of a target sentence \(y\) given the source sentence \(x\): \(p(y|x)\) . The encoder converts the input in the source language into a set of vectors while the decoder converts the set of vectors into the target language through an attention mechanism, one word at a time. The attention mechanism was introduced to keep track of context in longer sentences \cite{Bahdanau2014}.

The NMT model produces the translation sentence by generating one target word at every time step. Given an input sequence \(X=(x_1,...,x_{T_x})\) and previously translated words \((y_1,...,y_{i-1})\), the probability of the next word \(y_i\) is 
\begin{equation}
p(y_i|y_1,...,y_{i-1},X) = g(y_{i-1},s_i,c_i)
\end{equation}
where \(s_i\) is the decoder hidden state for time step \(i\) and is computed as
\begin{equation}
s_i = f(s_{i-1},y_{i-1},c_i).
\end{equation}

Here, \(f \) and \(g\) are nonlinear transform functions, which can be implemented as long short-term memory (LSTM) network \cite{Hochreiter1997} or gated recurrent units (GRU) \cite{Cho2014} in recurrent neural machine translation (RNMT), and \(c_i\) is a distinct context vector at time step \(i\), which is calculated as a weighted sum of the input annotations \(h_j\)
\begin{equation}
\sum_{j=1}^{T_x} a_{i,j}h_j
\end{equation}
where \(h_j\) is the annotation of \(x_j\) calculated by a bidirectional Recurrent Neural Network. The weight \(a_{i,j}\) for \(h_j\) is calculated as
\begin{equation}
a_{i,j} = \frac{\exp{e_{i,j}}}{\sum_{t=1}^{T_x} \exp{e_{i,t}}}
\end{equation}
and
\begin{equation}
e_{i,j} = v_a\tanh(Ws_{i-1}+Uh_j)
\end{equation}
where \(v_a\) is the weight vector, \(W\) and \(U\) are the weight matrices.

All of the parameters in the NMT model, represented as $\theta$, are optimized to maximize the following conditional log-likelihood of the \(M\) sentence aligned bilingual samples
\begin{equation}
L(\theta) = \frac{1}{M}\sum_{m=1}^{M} \sum_{i=1}^{T_y}\log(p(y_i^m|y_{<i}^m, X^m,\theta))
\end{equation}

To remove the recurrence and enable parallelization across multiple GPUs during training, the convolutional neural networks were used to create the convolutional NMT (CNMT) encoder-decoder architecture \cite{Gehring2017,Wu2019}. The CNMT utilizes 1-dimensional convolutional layers followed by gated linear units, GLU \cite{Dauphin2017}. The decoders compute and apply attention to each of the layers. The model uses positional embeddings along with residual connections \cite{Gehring2017}.

The transformer \cite{Vaswani2017,Dehghani2019} architecture was introduced to remove the recurrence and convolutions of previous architectures. The transformer is based solely on multi-headed self-attention layers. It enables parallelization across multiple GPUs, thereby, reducing training time. The architecture is used in current state-of-the-art translation systems \cite{Edunov2018,Ott2018}.

In this work, we used a unidirectional LSTM encoder-decoder architecture with Luong attention \cite{Luong2015}. This is a simple recurrent neural network RNMT architecture. Our approach is not architecture-dependent and can be applied to the other architectures or other more enhanced implementations of the RNMT.

\subsection{Leveraging Monolingual Data for NMT}
\label{lit-lev}
The use of monolingual data of the target and/or source language has been studied extensively to improve the performance of neural translation models, especially in low resource settings. \cite{Gulcehre2017} explored integrating language models trained on monolingual data into NMT systems, \cite{Currey2017,Burlot2018} proposed augmenting a copy or slightly modified copy respectively of the target data as source, \cite{Sennrich2016a} proposed the back-translation approach, \cite{Zhang2016} proposed the forward translation and \cite{He:2016:DLM:3157096.3157188} used both forward and back-translations to improve the translation models. The back-translation approach has been shown to outperform other approaches in low and high resource languages \cite{Edunov2018,Hoang2018}.

Various studies have investigated back-translation to improve the backward model, to select the most suitable generation/decoding methods for generating the synthetic data and to reduce the impact of higher ratio of the synthetic to the authentic bitext. The quality of the models trained using back-translation depends on the quality of the backward model \cite{Edunov2018,Fadaee2018,Hoang2018,Burlot2018,Graca2019,Kocmi2019,Yang2019}. To improve the quality of the synthetic parallel data, \cite{Hoang2018} used iterative back-translation – iteratively using the back-translated data to improve both the backward and forward models. \cite{Kocmi2019} and \cite{Dabre2019} used high resource languages through transfer learning and \cite{Zhang2018} explored the use of both target and source monolingual data to improve both the backward and forward models. \cite{Niu2018} trained a bilingual system based on \cite{Johnson2017} to do both forward and backward translations, eliminating the need for two separate models. \cite{Poncelas2019} used Transductive data selection methods to select monolingual data that are in the same domain as the test set for back-translation, improving performance.

The works of \cite{Fadaee2018,Poncelas2018} have found that the ratio of synthetic to authentic data affects the performance of the models most. When the ratio is high, the model tends to learn more from the synthetic data, which contains more mistakes than the authentic data. Investigations have found that the sampling approach of synthetic data generation and adding noise to beam search output outperforms the regular beam decoding technique \cite{Edunov2018,Imamura2018}. These approaches were said to improve the models by enhancing source-side diversity. \cite{Caswell2019} claimed, instead, that the noise only indicates to the model that the input is either synthetic or authentic, enabling the model to better utilize the two data. \cite{Yang2019} and \cite{Caswell2019} used tags (and gates) to enable the model to distinguish between the data and the approach has been shown to efficiently utilize more synthetic data, outperforming standard back-translation and enhancing the efficiency of iterative back-translation.

\subsection{Domain Adaptation}
\label{lit-pret}
Domain adaptation is the use of a usually few amount of in-domain data to improve the performance of an out-of-domain (general purpose) model before deployment. The amount of the in-domain parallel data is usually not sufficient to train a very good model and the general purpose models usually performs poorly \cite{Koehn2017a}. There are two categories of domain adaptation -- \emph{data centric} and \emph{model centric} \cite{Chu2018} with each having several techniques. The techniques in these classifications include using monolingual data \cite{Gulcehre2017}, synthetic data generation \cite{Sennrich2016a}, using data selection \cite{VanderWees2017} and using tagged out-of-domain parallel data \cite{Chu2017} and fine-tuning \cite{Sennrich2016a}

Pre-training has been used successfully in various machine learning tasks to improve performance when the data is not enough to train a good enough model. It was used for training word embeddings \cite{Mikolov2013}, in computer vision \cite{DBLP:journals/corr/abs-1902-11128}, fine-tuning NMT models \cite{Edunov2018} and as transfer learning in low resource NMT \cite{Zoph2016,Kocmi2018}. The transfer learning for machine translation approach involves training a model on a high resource language pair and transferring the training on a low resource pair. The works of \cite{Zoph2016,Nguyen2017,Kocmi2018} have shown tremendous improvements over models that are trained with the low resource data from scratch.

\begin{figure}[t!]
\centering
\includegraphics[clip, trim=6.5cm 18.5cm 7.5cm 3.0cm, width=0.75\textwidth]{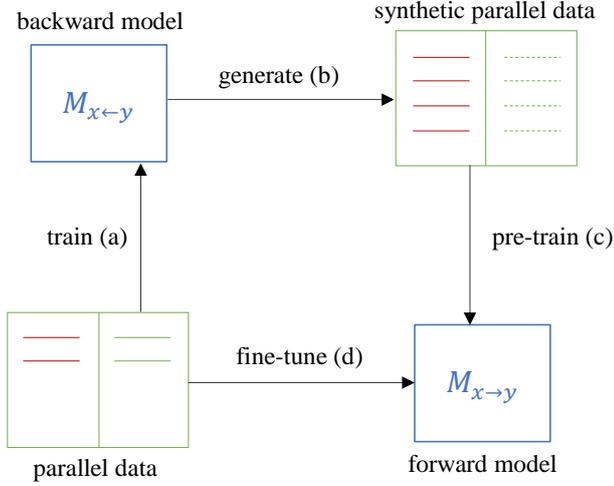}
\caption{Tag-less Back-Translation: Training the forward model on the synthetic parallel data generated using the backward model. The forward model is then fine-tuned on the authentic data.}
\label{fig:1}
\end{figure}

In back-translation, \cite{Sennrich2016a} showed that fine-tuning a pre-trained model on in-domain data improves the quality of back-translated model. \cite{Popel2018} pre-trained the model on the authentic data and fine-tunes it on the mixed synthetic and authentic data. \cite{Kocmi2019} and \cite{Dabre2019} pre-trained a model on a high resource language and fine-tunes it on a low resource language pair.

\section{The Proposed Method}
\label{method}
The approach is shown in Fig. 1. As illustrated in Algorithm 1, the authentic parallel data: \(D^P =\{(x^{(u)}, y^{(u)})\}_{u=1}^U\) is used to train a target-to-source model, \(M_{x \leftarrow y}\). This model -- the \emph{backward} model -- is used to translate the monolingual target data, \(Y =\{(y^{(v)})\}_{v=1}^V\), to generate the synthetic parallel data: \(D^\prime =\{(x^{(v)}, y^{(v)})\}_{v=1}^V\). Instead of mixing the two data to train a forward (target) model, we used only the synthetic data to pre-train the forward model, \(M_{x \rightarrow y}\), until no improvement is observed on the development set. Finally, the forward model is fine-tuned on authentic data.

It was shown in \cite{Kocmi2019} that using different vocabulary each during pre-training and fine-tuning leads to drop in performance because, it was said, independent vocabulary use different identifiers even for the same subwords and the network loses benefits of the weights learned during pre-training. The authors proposed learning a joint BPE on a mixture of both the pre-training and fine-tuning data and this has been shown to achieve better results in domain adaptation. In this approach, we have access to both the out-of-domain (synthetic) and the in-domain (authentic) parallel data. This, therefore, enables us to learn a joint BPE and build the training vocabulary for both pre-training and fine-tuning.

\begin{table}[ht]
\label{tab:algorithm}
\renewcommand{\arraystretch}{1.3}
\begin{tabular}{l}
\textbf{Algorithm 1:} Tag-less Back-Translation \\
\noalign{\smallskip}
\hline\noalign{\smallskip}
\textbf{Input:} Parallel data \(D^P =\{(x^{(u)}, y^{(u)})\}_{u=1}^U\) and \\
\quad \quad \quad \quad Monolingual target data \(Y =\{(y^{(v)})\}_{v=1}^V\) \\
\noalign{\smallskip}\hline\noalign{\smallskip}
1: \textbf{procedure} BACK-TRANSLATION \\
2: \quad Train backward model \(M_{x \leftarrow y}\) on bilingual data \(D^P\) \\
3: \quad Use \(M_{x \leftarrow y}\) to create \(D^\prime =\{(x^{(v)}, y^{(v)})\}_{v=1}^V\), for \(y \in Y\); \\
4: \quad Pre-train forward model \(M_{x \rightarrow y}\) on parallel data \(D^\prime\); \\
5: \quad Fine-tune the forward model \(M_{x \rightarrow y}\) on parallel data \(D^P\); \\
6: \textbf{end procedure} \\
\noalign{\smallskip}
\textbf{Output:}  forward model \(M_{x \rightarrow y}\) \\
\noalign{\smallskip}\hline
\end{tabular}
\end{table}

\section{Experiments}
\label{exp}

\subsection{Set-up}
\label{exp-setup}
We used the TensorFlow \cite{Abadi2015} implementation of the OpenNMT \cite{Klein2017} framework to train the models -- the NMTSmallV1 configuration. The set-up is based on the NMTSmallV1 configuration. Specifically, the configuration is a 2-layer unidirectional LSTM encoder-decoder model with Luong attention \cite{Luong2015}. It has 512 hidden units and a vocabulary size of 50,000 for both source and target languages. We used Adam \cite{Kingma2015} optimizer and a batch size of 64 with a dropout probability of 0.3, a static learning rate of 0.0002 and the models are evaluated on the development set after every 5,000 training steps. The models were evaluated using the bi-lingual evaluation understudy metric, BLEU \cite{Papineni2002}, specifically the multi-bleu \cite{Koehn:2007:MOS:1557769.1557821} implementation. The models are trained until there is no improvement of over 0.2 BLEU after four training steps. As stated in Section \ref{method}, the learning of BPE on the training data and the building of training vocabulary for both pre-training and fine-tuning was done on the mixture of the synthetic and authentic parallel data. During fine-tuning, we, therefore, only change the training data.

\begin{table}
\renewcommand{\arraystretch}{1.3}
\caption{Data Used}
\label{tab:1}
\begin{tabular}{cccccc}
\hline
\multirow{2}{*}{data} & \multicolumn{3}{c}{train} & \multirow{2}{*}{dev} & \multirow{2}{*}{test} \\ \cline{2-4}
 & sentences & \multicolumn{2}{c}{words (vocab)} &  &  \\ \hline
\multirow{2}{*}{} & \multirow{2}{*}{} & En & Vi & \multirow{2}{*}{} & \multirow{2}{*}{} \\ \cline{3-4}
 IWSLT’15 En-Vi & 133, 317 & \makecell[tc]{2,837,240\\ (50,045)} & \makecell[tc]{2,688,387\\ (103,796)} & 1, 553 & 1, 268 \\ \noalign{\smallskip}\hline\noalign{\smallskip}
\multirow{2}{*}{} & \multirow{2}{*}{} & En & De & \multirow{2}{*}{} & \multirow{2}{*}{} \\ \cline{3-4}
 IWSLT’14 En-De & 153, 348 & \makecell[tc]{2,706,255\\ (54,169)} & \makecell[tc]{3,311,508\\ (25,615)} & 6, 970 & 6, 750 \\ \noalign{\smallskip}\hline\noalign{\smallskip}
\begin{tabular}[c]{@{}c@{}}WMT’14 En-De –\\ Monolingual English\end{tabular} & 666,585 & \multicolumn{2}{c}{16,700,106 (344,138)} & - & - \\ \hline
\end{tabular}
\end{table}

\subsection{Data}
\label{exp-data}
For this work, we use the preprocessed low resource English-Vietnamese parallel data \cite{Luong2015} of the IWSLT 2015 Translation task \cite{Cettolo2017}. We used the 2012 and 2013 test sets for development and testing respectively. We also used the data from the IWSLT 2014 German-English shared translation task \cite{Cettolo2014} as the second language pair, pre-processed using the data clean-up as well as the train, development and test split in \cite{Ranzato2016}. For the monolingual data, we used the preprocessed English monolingual data of WMT 2014 English-German translation task \cite{Bojar2014-findings}. We shuffled the monolingual data and selected 666,585 monolingual sentences which is five times as much as the En-Vi parallel data. The statistics of the datasets are shown in Table \ref{tab:1}. We learned byte pair encoding (BPE) \cite{Sennrich2016b} with 10,000 merge operations on the training dataset and applied it on the train, development and test datasets. Afterwards, we build the vocabulary on the training dataset. For all the experiments, we used thrice as much of the monolingual as the available parallel data in both of the languages except when we experimented with the ratio of 1:5 (parallel to monolingual data) for the English-Vietnamese NMT.

\subsection{Models}
\label{models}
To compare the performance of our approach with that of the previous works, we implemented the following methods to train translation models on the English-Vietnamese and English-German NMT using the data presented in Section \ref{exp-data} above. All models were trained using the same settings stated in Section \ref{exp-setup}
\begin{itemize}
	\item We first train \emph{baseline} models using the available authentic parallel data only. In the models, the baselines have the English language as the target language -- Vi-En and De-En.
	\item We then train the backward models also on the authentic parallel data using English language as the source language -- En-Vi and En-De. The models are used for the generation of the additional synthetic parallel data for the back-translation approach.
	\item We implemented the various back-translation strategies namely standard back-translation -- \emph{standard\_bt}, the tagged back-translation -- \emph{tagged\_bt} and the tag-less back-translation -- \emph{tag-less\_bt (joint BPE)} using the authentic and synthetic parallel data.
\end{itemize}

\section{Results}
\label{result}
All scores reported are statistically significant with p \textless~0.05. We used the paired bootstrap resampling of \cite{koehn-2004-statistical} as implemented in \cite{NEURIPS2019_9015} to estimate the statistical significance confidence scores. See Table \ref{stat-sig} in Appendix 1 for confidence scores.

\subsection{English-Vietnamese Low Resource NMT}
\label{envi-nmt}

The evaluation scores of the best models and the improved models obtained after taking the checkpoint averaging of the last 8 checkpoints are shown in Table \ref{tab:2}. We first created a forward Vietnamese-English (Vi-En) model, \emph{baseline}, on the available authentic parallel data. The model trained for 75,000 steps before the stopping condition was met. The baseline model was trained further to $110,000$ training steps but the performance continued to flatten without observing any improvement. The model achieved the best single-checkpoint score of 21.19 BLEU at the 65,000$^{th}$. We then trained a backward (En-Vi) model, \emph{backward}. After the stoppage condition were met, after 55,000 training steps, the best performing single-checkpoint for the backward model achieved a BLEU score of 24.78 at the 50,000$^{th}$ training step. Averaging the last 8 checkpoints gave the best performance -- $25.79$ BLEU. This average model was used to back-translate the monolingual English data to generate synthetic parallel data. We mixed the two data -- synthetic and authentic -- without differentiating between the two and used the resulting large dataset to train a forward model. We labelled this model as \emph{standard\_bt} – for standard back-translation \cite{Sennrich2016a}. This model was trained for 165,000 before the stopping condition were met. It achieved a single-checkpoint best BLEU score of 24.46 at the 155,000$^{th}$ training step.

\begin{table}
\caption{Performance of the Tag-less Back-translated model compared to the baseline and standard back-translation models for Vietnamese-English and German-English translations. Evaluation scores on the test set. The tag-less approaches show results of pre-training and fine-tuning.}
\label{tab:2}
\renewcommand{\arraystretch}{1.3}
\begin{tabular}{llllll}
\hline\noalign{\smallskip}
data & & \emph{baseline} & \emph{standard\_bt} & \emph{tag-less\_bt} & \makecell[l]{\emph{tag-less\_bt}\\(joint BPE)} \\
\noalign{\smallskip}\hline\noalign{\smallskip}
\multirow{2}{*}{Vi-En} & \makecell[l]{best checkpoint \\ BLEU (training \\ step)} & 21.19 (65k) & 24.46 (155k) & \makecell[l]{17.85 (90k) \\ 25.16 (145k)} & \makecell[l]{18.60 (105k) \\ 26.53 (165k)}\\ \noalign{\smallskip}\noalign{\smallskip}
 & average & 22.22 & 25.28 & \makecell[l]{17.96 \\ 25.77} & \makecell[l]{18.59 \\ 26.83} \\ \noalign{\smallskip}\hline \noalign{\smallskip}
\multirow{2}{*}{De-En} & \makecell[l]{best checkpoint \\ BLEU (training \\ step)} & 20.30 (75k) & 25.11 (150k) & \makecell[l]{3.01 (75k) \\ 25.16 (120k)} & \makecell[l]{5.43 (60k) \\ 28.31 (155k)}\\ \noalign{\smallskip}\noalign{\smallskip}
 & average & 20.95 & 25.87 & \makecell[l]{3.03 \\ 26.03} & \makecell[l]{5.13 \\ 28.83} \\
\noalign{\smallskip}\hline
\end{tabular}
\end{table}

We mixed the synthetic and authentic parallel data and learned a joint BPE on the resulting training dataset and build the training vocabulary. We applied the BPE on the synthetic data for pre-training and on the authentic data for fine-tuning. We trained a model, labelled \emph{tag-less\_bt (joint BPE)}, using this approach. The model achieved a single-checkpoint score of 18.60 BLEU during pre-training and improved to 26.53 BLEU after fine-tuning. The average fine-tuned model was better by about 0.30 BLEU. The average pre-trained model performed very low compared to the baseline and the standard back-translation models -- 18.59 BLEU vs 22.22 and 25.28 BLEUs respectively. This is obviously because the quality of the data used in the training the model -- the synthetic data -- is lower than that of the other two. The quality of the synthetic data, although generated from a reasonably good backward model, is still not sufficient to train a model whose quality can compare to the other models that are trained in whole or in part with the authentic data. Fine-tuning the model on the authentic data results in a sharp rise in performance. The model was fine-tuned until the stopping condition was met. The approach outperformed the baseline and standard back translation models by 5.34 and 2.07 BLEUs respectively. The gap in performance was, however, reduced to 4.61 and 1.55 BLEUs after checkpoint averaging.

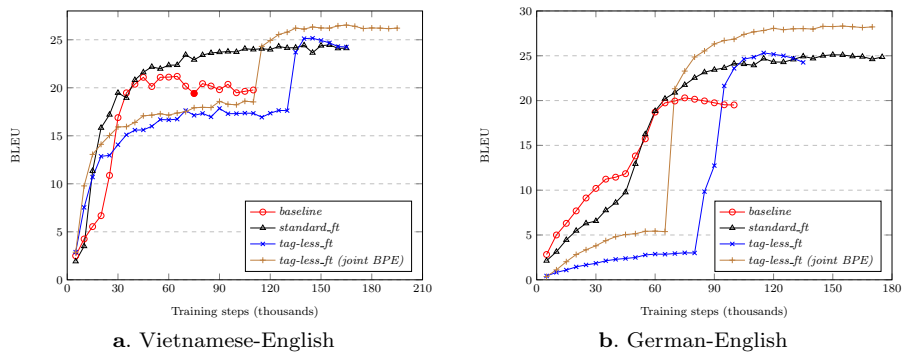
\begin{figure}[t!]
\centering
\begin{minipage}[t]{0.48\textwidth}
\resizebox{1.0\linewidth}{!}{%
\begin{tikzpicture}
\begin{axis}[
    xlabel={Training steps (thousands)},
    ylabel={BLEU},
    xmin=0, xmax=210,
    ymin=0, ymax=28,
    xtick={0,30,60,90,120,150,180,210},
    ytick={0,5,10,15,20,25,30},
    legend pos=south east,
    legend cell align={left},
    ymajorgrids=true,
    grid style=dashed,
    width=10cm,
    height=8cm,
]

\addplot[
    color=red,
    mark=o,
    ]
    table {envibaseline.txt};
    \addlegendentry{\emph{baseline}}

\addplot[
    color=black,
    mark=triangle,
    ]
    table {envistandardbt.txt};
    \addlegendentry{\emph{standard\_ft}}
    
\addplot[
    color=blue,
    mark=x,
    ]
    table {envitagless.txt};
    \addlegendentry{\emph{tag-less\_ft}}

\addplot[
    color=brown,
    mark=+,
    ]
    table {envitaglessjointbpe.txt};
    \addlegendentry{\emph{tag-less\_ft (joint BPE)}}

\filldraw[red] (75,194.1) circle (2pt) node[anchor=west]{};

\end{axis}
\end{tikzpicture}
}
\centering
\textbf{a}. Vietnamese-English
\end{minipage}%
\hfill
\begin{minipage}[t]{0.48\textwidth}
\resizebox{1.0\linewidth}{!}{%
\raggedleft
\begin{tikzpicture}
\begin{axis}[
    xlabel={Training steps (thousands)},
    ylabel={BLEU},
    xmin=0, xmax=180,
    ymin=0, ymax=30,
    xtick={0,30,60,90,120,150,180},
    ytick={0,5,10,15,20,25,30},
    legend pos=south east,
    legend cell align={left},
    ymajorgrids=true,
    grid style=dashed,
    width=10cm,
    height=8cm,
]

\addplot[
    color=red,
    mark=o,
    ]
    table {endebaseline.txt};
    \addlegendentry{\emph{baseline}}

\addplot[
    color=black,
    mark=triangle,
    ]
    table {endestandardbt.txt};
    \addlegendentry{\emph{standard\_ft}}
    
\addplot[
    color=blue,
    mark=x,
    ]
    table {endetagless.txt};
    \addlegendentry{\emph{tag-less\_ft}}

\addplot[
    color=brown,
    mark=+,
    ]
    table {endetaglessjointbpe.txt};
    \addlegendentry{\emph{tag-less\_ft (joint BPE)}}

\end{axis}
\end{tikzpicture}
}
\centering
\textbf{b}. German-English
\end{minipage}
\caption{Tag-less back-translation: pre-training on synthetic data and fine-tuning on authentic data. Showing how this technique compares to the baseline and the standard back-translation approaches on the test set.}
\label{fig:2}
\end{figure}

We experimented the other pre-train and fine-tune approach, learning the BPE only on the synthetic data. We build the vocabulary on the synthetic training data after applying the BPE. The synthetic corpus was used to pre-train a forward model for 130,000 steps, achieving a single-checkpoint best score of 17.85 BLEU. The authentic parallel data was then used to fine-tune the model for a further 35,000 training steps. Stopping at each of these steps were based on the stopping condition. The performance of the \emph{tag-less\_bt} model improved to 25.16 BLEU after fine-tuning. Although this approach was shown to outperform the baseline and standard back-translation, it underperformed the joint BPE implementation of the tag-less approach by 1.06 BLEU. In Figures \ref{fig:2}a and \ref{fig:2}b, we show how the BLEU scores continue to improve with increase in the training steps. The model trained using the \emph{tag-less\_bt (joint BPE)} approach continued to outperform the three others after fine-tuning.

\subsection{English-German Low Resource NMT}
\label{ende-nmt}
We conducted the same set of experiments presented in section \ref{envi-nmt} on the second low resource dataset, the English-German IWSLT'14 parallel dataset. This data, as presented in Table \ref{tab:1}, is made up of a little bit more than 150,000 parallel sentences. We first trained a backward (En-De) model on the available parallel data. This model maxed-out performance on the test set, based on the set-up, at 10.25 BLEU after averaging the last 8 checkpoints. It stopped training at the 80,000$^{th}$ training steps and achieving the best single model performance at the 65,000$^{th}$ -- 10.03 BLEU. We used the average model to generate the synthetic data, translating the available English monolingual data. We trained four separate forward (De-En) models based on the approaches we explained earlier. The first is the baseline trained on the available authentic data, the \emph{standard\_bt} on the mixture of the authentic and synthetic data without differentiating, the \emph{tag-less\_bt} pre-trained on the synthetic data and fine-tuned on the authentic data having learned the BPE on the synthetic data and updating the vocabulary before fine-tuning and, finally, the \emph{tag-less\_bt (joint BPE)} trained also using the tag-less approach but having learned the BPE and built the vocabulary on the mixture of the synthetic and authentic data.

The results of evaluating the models after training using the various approaches are presented in Table \ref{tab:2} The baseline achieved a modest average performance of 20.95 BLEU after training for 100,000 training steps. The performance on the dataset was improved after applying standard back-translation, achieving a huge +4.92 BLEU improvement over the baseline. The tag-less approach, though better, did not achieve a huge improvement over back-translation (only +0.16 BLEU) but after applying the improved tag-less (joint BPE), as shown in the previous section, we achieved huge +2.96 BLEU increase in performance. This +2.8 and +7.88 BLEUs over the previous tag-less approach and the baseline respectively.

For all the subsequent experiments, unless stated otherwise, we used the joint BPE technique to implement the tag-less back-translation approach as it is shown to be the most successful variant.

\begin{figure}[t!]
\centering
\begin{minipage}[t]{0.48\textwidth}
\resizebox{1.0\linewidth}{!}{%
\begin{tikzpicture}
\begin{axis}[
    xlabel={Training steps (thousands)},
    ylabel={BLEU},
    xmin=0, xmax=200,
    ymin=0, ymax=30,
    xtick={0,30,60,90,120,150,180},
    ytick={0,5,10,15,20,25,30},
    legend pos=south east,
    legend cell align={left},
    ymajorgrids=true,
    grid style=dashed,
    width=10cm,
    height=8cm,
]

\addplot[
    color=blue,
    mark=o
    ]
    table {envitaglessjointbpe.txt};
    \addlegendentry{\emph{tag-less bt (joint BPE)}}

\addplot[
    color=red,
    mark=x
    ]
    table {envitagged.txt};
    \addlegendentry{\emph{tagged bt}}
    
\filldraw[red] (125,240.8) circle (2pt) node[anchor=west]{};

\end{axis}
\end{tikzpicture}
}
\centering
\textbf{a}. Vietnamese-English
\end{minipage}%
\hfill
\begin{minipage}[t]{0.48\textwidth}
\resizebox{1.0\linewidth}{!}{%
\raggedleft
\begin{tikzpicture}
\begin{axis}[
    xlabel={Training steps (thousands)},
    ylabel={BLEU},
    xmin=0, xmax=260,
    ymin=0, ymax=30,
    xtick={0,30,60,90,120,150,180,210,240},
    ytick={0,5,10,15,20,25,30},
    legend pos=south east,
    legend cell align={left},
    ymajorgrids=true,
    grid style=dashed,
    width=10cm,
    height=8cm,
]

\addplot[
    color=blue,
    mark=o
    ]
    table {endetaglessjointbpe.txt};
    \addlegendentry{\emph{tag-less bt (joint BPE)}}

\addplot[
    color=red,
    mark=x
    ]
    table {endetagged.txt};
    \addlegendentry{\emph{tagged bt}}

\end{axis}
\end{tikzpicture}
}
\centering
\textbf{b}. German-English
\end{minipage}
\caption{Tagged Vs Tag-less back-translation.}
\label{fig:3}
\end{figure}
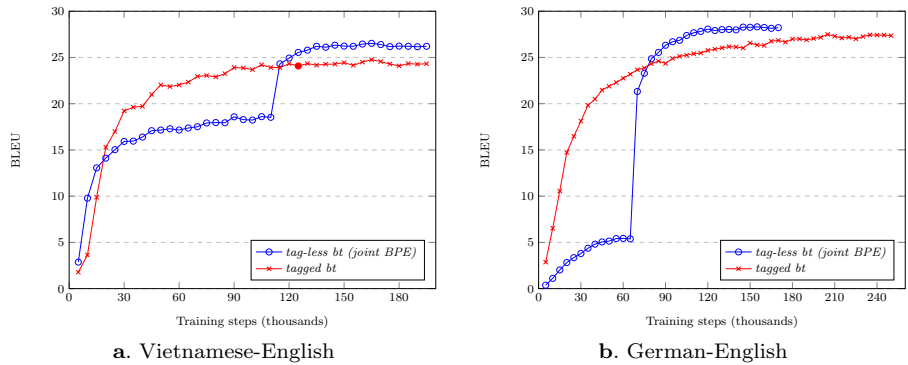

\subsection{Tagged Vs Tag-Less Back-translation}
\label{result-tagged}
We compared the performance of the \emph{tag-less\_bt} model -- our technique -- with that of the successful tagged back-translation of \cite{Caswell2019} on the English-Vietnamese data. The synthetic sources were labelled with the \emph{$<$BT$>$} token at the beginning of each sentence and mixed with the authentic sources to generate the mixed tagged parallel corpus. This mixed data is used to train the forward tagged back-translation model -- \emph{tagged\_bt}. The \emph{tagged\_bt} model stopped at 125,000 steps and the training was continued up to 195,000 steps to equal the number of training steps reached by the \emph{tag-less\_bt} model.  While the tagged approach underperformed the best score of our technique by 1.78 BLEU, it was able to outperform the single-checkpoint standard back-translation by 24.76 to 24.46 BLEUs respectively (+0.3 BLEU) but underperformed the average standard back-translation model by 0.23 BLEU.

\begin{table}
\caption{Performance of Tag-less Back-translation compared to the Tagged back-translation model for Vietnamese-English translation. Evaluation scores on the test set.}
\label{tab:3}
\renewcommand{\arraystretch}{1.3}
\begin{tabular}{llll}
\hline\noalign{\smallskip}
 && \emph{tagged\_bt} &  \makecell[l]{\emph{tag-less\_bt (joint BPE)}} \\
\noalign{\smallskip}\hline\noalign{\smallskip}
\multirow{2}{*}{En-Vi}&\makecell[tl]{best checkpoint BLEU\\ (training step)} & 24.76 (165k) & 26.53 (165k) \\
&average & 25.05 & 26.83 \\ \noalign{\smallskip}\hline\noalign{\smallskip}
\multirow{2}{*}{En-De}&\makecell[tl]{best checkpoint BLEU\\ (training step)} & 27.49 (205k) & 28.31 (155k) \\
&average & 27.75 & 28.83 \\ \noalign{\smallskip}\hline\noalign{\smallskip}
\end{tabular}
\end{table}

\begin{table}
\caption{Performance of Tag-less Back-translation on the test set: pre-training on synthetic data and fine-tuning on authentic data Vs pre-training on authentic data and fine-tuning on synthetic data for Vietnamese-English.}
\label{tab:4}
\renewcommand{\arraystretch}{1.3}
\begin{tabular}{lll}
\hline\noalign{\smallskip}
 & \emph{tag-less\_bt} & \emph{reverse\_tag-less\_bt} \\
\noalign{\smallskip}\hline\noalign{\smallskip}
\makecell[l]{best checkpoint BLEU \\ (training step)} & 25.16 (145k) & \makecell[l]{21.19 (65k) - pre-train \\ 18.95 (100k) - fine-tune} \\
\noalign{\smallskip}\noalign{\smallskip}
average & 25.77 & 18.91 \\
\noalign{\smallskip}\hline
\end{tabular}
\end{table}

Finally, we trained a forward model using the tagged back-translation for English-German NMT to compare the performance with our approach on this data. The tagged approach took a further 50,000 training steps to reach a single model best of 27.49 BLEU, but still underperforming the tag-less approach by 0.82 BLEU. The best model obtained after averaging checkpoints was also achieved using our approach, a performance of 28.83 BLEU compared to the tagged 27.75 BLEU, an improvement of 1.08 BLEU. The performances of these models, evaluated on the test set is shown in Table \ref{tab:3}. On this data, the tagged approach performed better than the standard back-translation by +1.88 BLEU on the average models. It can be seen in Figures \ref{fig:2} and \ref{fig:3} that in both of the experiments conducted on the two data, our tag-less approach out-performed the rest of the back-translation approaches.

This supports the hypothesis that although the tagged back-translation involves explicit differentiating between the two data using tags, the model trained on the approach may not be able to differentiate between them completely during training as observed in the mixed performance of the models trained on the two different data.

\subsection{Fine-tuning: Synthetic Vs Authentic Data}
\label{result-finet}
Our technique proposed pre-training the forward model on the synthetic parallel data and fine-tuning the model afterwards on the authentic data. This was proposed to enable the model to unlearn the mistakes it learned from the synthetic data using correct sentences in the authentic parallel data. We experiment the other way round to investigate the effects of pre-training on the authentic data and fine-tuning on the synthetic data. We used the baseline as the pre-trained model and fine-tune it on the synthetic data. This approach was labelled as \emph{reverse\_tag-less\_bt}. This approach did not show any benefit to the final forward model, see Fig. 3. As expected, the performance of the model decreased and the curve flattens as the number of training steps increases. The best and average scores are shown in Table 3.

\begin{figure}[t!]
\centering
\begin{tikzpicture}
\begin{axis}[
    xlabel={Training steps (thousands)},
    ylabel={BLEU},
    xmin=0, xmax=180,
    ymin=0, ymax=28,
    xtick={0,30,60,90,120,150,180},
    ytick={0,5,10,15,20,25,30},
    legend pos=south east,
    legend cell align={left},
    ymajorgrids=true,
    grid style=dashed,
]
 
\addplot[
    color=red,
    mark=o
    ]
    table {envibaselinecompare.txt};
    \addlegendentry{\emph{baseline}}

\addplot[
    color=black,
    mark=x
    ]
    table {envifinetunebaseline.txt};
    \addlegendentry{\emph{finetune}}

\end{axis}
\end{tikzpicture}
\caption{Fine-tuning the baseline model on the synthetic data. Evaluation scores on the test set.}
\label{fig:4}
\end{figure}
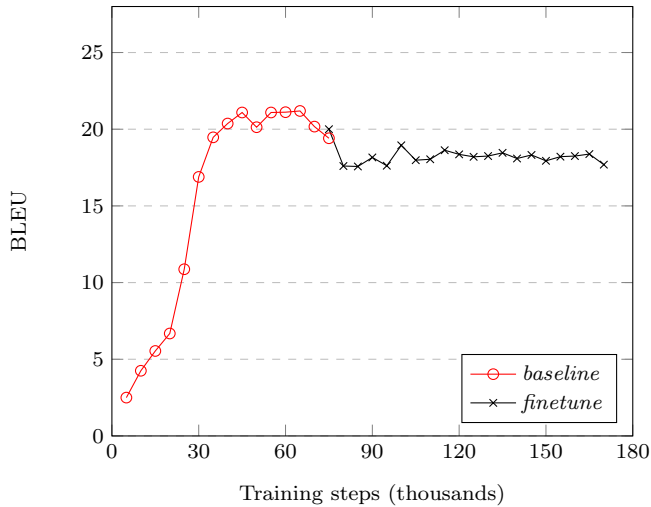

\subsection{Quantity of Monolingual Data}
\label{result-quantity}
As stated earlier, it was found that as the more synthetic data increases, a point is reached where the performance starts to deteriorate \cite{Fadaee2018}. Instead, our work hypothesizes that the performance of the model will start to decrease only if it is not able to differentiate between the synthetic and authentic training data and, therefore, efficiently learning from the two. We also pointed out that since the data is mixed in both the standard and tagged back-translation approaches, the model may not be able to completely differentiate between the data, although in the latter approach, the model is expected to treat the tagged synthetic sources as a different domain. We, therefore, experiment with different ratios of the authentic to synthetic data to verify this claim. We sample the authentic to synthetic data in the ratios, 1:1, 1:3 and 1:5. The results are shown in Table \ref{tab:5}.

In the tagged approach, the single-checkpoint best scores continue to rise from using the same amount of monolingual data for back-translation to using three times the authentic data of the monolingual data for back-translation. But, as observed, the performance dropped slightly when we used five times the amount of available parallel data. However, the performance of the tag-less back-translation models continues to increase steadily when the ratio of authentic to monolingual data is increased. We observed the performance to improve by about 0.25 BLEU when the amount of synthetic data is tripled and doubled to about 0.5 BLEU after adding another double amount of the synthetic data to the training data. It can also be observed that there was a very low improvement over the performance of \emph{baseline} and serious underperformance compared to the tag-less approach when we used the same amount of synthetic data with the authentic data to train the models -- 22.73 BLEU vs 22.22 and 26.15 BLEUs respectively. Overall, we obtained a 3.42, 1.78 and 1.67 BLEU improvements on the average models using the tag-less approach over the tagged approach on the ratios experimented respectively.

\begin{table}
\caption{Using different ratios of the authentic to synthetic data for Vietnamese-English translation. Evaluation scores of the models on the test set.}
\label{tab:5}
\renewcommand{\arraystretch}{1.3}
\begin{tabular}{llllllll}
\hline\noalign{\smallskip}
 & \multicolumn{3}{c}{\emph{tagged\_bt}}& & \multicolumn{3}{c}{\emph{tag-less\_bt} (joint BPE)} \\
\noalign{\smallskip}
 & \textbf{1:1} & \textbf{1:3} & \textbf{1:5} & & \textbf{1:1} & \textbf{1:3} & \textbf{1:5} \\
\noalign{\smallskip}\hline\noalign{\smallskip}
\makecell[tl]{best checkpoint BLEU\\ (training step)} & \makecell[tl]{22.73 \\ (85k)} & \makecell[tl]{24.76 \\ (165k)} & \makecell[tl]{24.62 \\ (155k)} & & \makecell[tl]{26.29 \\ (130k)} & \makecell[tl]{26.53 \\ (165k)} & \makecell[tl]{\textbf{26.91} \\ (185k)} \\
\noalign{\smallskip}\noalign{\smallskip}
average & 22.73 & 25.05 & 25.47 & & 26.15 & 26.83 & \textbf{27.14} \\
\noalign{\smallskip}\hline
\end{tabular}
\end{table}

It can be observed also that the performance of the model trained on the 1:1 ratio of monolingual to synthetic data using our approach is very good compared to the model trained using the same amount of data in the tagged approach and subsequent increase in training data leads to steady improvements that at 1:5 ratio, the performance was improved by about 1 BLEU. This steady improvements can show that the model learned useful knowledge on the authentic data but only used the synthetic data for further improvements. Following this trend, we can, therefore, conclude that with more synthetic data compared to the authentic data, the model will only continue to learn and increase its performance if useful representations are learnable on the synthetic data.

\begin{table}
\caption{Before and after fine-tuning the English-Vietnamese standard and tagged back-translation NMT models on the authentic data. Evaluation scores on the test set.}
\label{tab:6}
\begin{tabular}{llllll}
\hline\noalign{\smallskip}
 & \multicolumn{2}{c}{\emph{tagged\_bt (1:5)}}& & \multicolumn{2}{c}{\emph{standard\_bt (1:3)}} \\
\noalign{\smallskip}
 & before & after& & before & after \\
\noalign{\smallskip}\hline\noalign{\smallskip}
\makecell[tl]{best checkpoint BLEU\\ (training step)} & \makecell[tl]{24.62 \\ (155k)} & \makecell[tl]{25.55 \\ (175k)}& & \makecell[tl]{24.46 \\ (155k)} & \makecell[tl]{25.15 \\ (180k)} \\
\noalign{\smallskip}\noalign{\smallskip}
average & 25.47 & 25.64& & 25.28 & 25.32 \\
\noalign{\smallskip}\hline
\end{tabular}
\end{table}

\subsection{Fine-tuning Standard And Tagged Back-Translations}
\label{result-finetune-others}
The work of \cite{Popel2018} reported no observable advantage of using the authentic data to train the forward model and fine-tuning it henceforth on the mixed data. Instead, we experimented training the forward model on the mixed data first and then fine-tune it on the authentic data. As shown in Table 6, this approach reaches the same performance as the old tag-less approach -- 25.15 BLEU -- using the same amount of synthetic sentences albeit after 30,000 more training steps but sill underperforming the joint BPE tag-less back-translation's 26.53 BLEU although training for additional 15,000 training steps. The better joint BPE tag-less approach converges earlier than fine-tuning the standard back-translation model, at 165,000 training steps.

We also explored the use of fine-tuning to determine whether or not the tagged approach will be able to cover the difference in performance with the tag-less approach. After fine-tuning the \emph{tagged\_bt (1:5)} model for a further 35,000 training steps, the performance gained was a significant +0.93 BLEU and only 0.17 BLEU over the average after just 20,000 steps of fine-tuning. The performance was still short of the \emph{tag-less\_bt (joint BPE)} (1:5) by a significant 1.36 BLEU.

\section{Discussion}
\label{discuss}
In this work, we proposed an approach for training the forward model in back-translation without using tags or noising the synthetic data. Translation models that are trained on the synthetic and authentic data have been shown to perform better when they are able to differentiate between the two data. Previous approaches have relied on the use of noise in back-translation \cite{Edunov2018} especially on low resource languages to improve the performance of models. The authors thought that the approach ensures source-side diversity which has been shown to benefit the models \cite{Imamura2018}. The approach was found out to only indicate to the forward model that the noised data is synthetic, enabling it to treat the data differently from the authentic data \cite{Caswell2019}. The use of tags has been shown to improve the performance of such models. In this work, we eliminated the need of using of the tags and showed that although it was successful at improving the performance -- proving it successful at indicating to the model that a data is synthetic and not authentic -- domain adaptation methods are more capable of ensuring the model differentiate between the data. The ability for the model to separate between the data is even more important in low resource languages where the available data is not enough to train standard backward model, thus generating synthetic data with a lot of noises.

Domain adaptation techniques techniques in machine translation ensures that a better model is trained, leveraging on a larger parallel data of either the same language pair but in a different domain -- fine-tuning --  or a different language pair -- transfer learning. In this technique, the two data are not tagged, mixed and left to the model to differentiate between them. They, rather, are used at the different stages of the training and this ensures the model performs in the target domain as expected. We utilized the synthetic data -- which is bigger but more prone to translation noises -- as the generic domain and the authentic data -- smaller but having more quality -- as the in-domain. This selection was not done until the reversed approach was shown not give the desired performance. The superiority of the approach over the successful tagging was shown through experimental results conducted on two low resource language pair: English-Vietnamese and English-German. In each of the languages considered, we obtained an improvement of more than 1 BLEU points over the tagged approach that outperformed the baseline and standard back-translation models.

We also test the performance of our technique when the amount of monolingual data is increased. We used different ratios of the authentic parallel to monolingual data used. We found that our technique was not only able to handle the increase in the synthetic data, but was able to attain rapid improvement given the smallest amount of synthetic data. We obtained a superior performance by a whopping 3.56 BLEU using the tag-less approach over the tagged approach when the amount of monolingual data is the same as the authentic data. The performance continued to steady increase as the amount of monolingual data is increased. The tagged approach could only handle tripling the amount of synthetic data but the performance started to decrease when the synthetic data was increased further. Using the same amount of synthetic data in ratios 1:1, 1:3 and 1:5, our technique outperformed the tagging technique by 3.42, 1.78 and 1.67 respectively (see average scores in Table \ref{tab:5}).

Our approach also provides one with the flexibility of using state-of-the-art domain adaptation methods to improve the performance of the already successful back-translation approach. Techniques such as using different dropout and/or learning rate during pre-training and fine-tuning may improve the performance of the forward model. The method may also be applied in high-resource languages since both of these settings -- low and high resource -- can benefit from the ability of the forward model to differentiate between synthetic and authentic data.

\section{Conclusions and Future Work}
\label{conclusion}
This work has shown that an NMT model pre-trained on synthetic data and fine-tuned on the authentic data outperforms the rather successful method of tagging the synthetic data in low resource NMT by enabling the forward model to differentiate between the authentic and synthetic training data. The approach, however, does not improve the performance when it is reversed and the forward model is pre-trained on the authentic data and then fine-tuned on the synthetic data. As expected, the reverse approach makes the model to unlearn the useful representations learned in favour of the noise in the synthetic data. This justifies our hypothesis that without differentiating between the two data, the synthetic data is most likely to hurt the performance of the forward model.

It was shown also, in this work, that the more synthetic data used, the better the performance of the forward model, though the most effective ratio was not yet determined through thorough experimentation. This will inform the basis of future works. We experiment fine-tuning the models trained using the standard and tagged back-translation approaches. Experimental results showed the standard back-translation equalling the performance of a variant of the tag-less approach after many more rounds of training. The performance of the tagged approach improved considerably but still trailed the tag-less approach. The most successful of the tag-less approach has been the one that involves learning a joint BPE and building the training vocabulary on the mixture of the synthetic and authentic parallel data. This approach is made possible, unlike in other fine-tuning conditions, because both the generic (synthetic) data and the in-domain (authentic) data are available during the process.

For future work, the use of different settings -- such as increasing or decreasing the learning rate, using dropout and L2 regularization, which may reduce overfitting on the in-domain (authentic) data as shown to be a likely problem in domain adaptation by \cite{Barone2017} -- for the pre-training and fine-tuning approaches can be explored to maximize the benefits of the domain adaptation approach in back-translation. The approach can also investigated to improve the forward translation approach -- which also leverages on the synthetic data for additional training data. Finally, we intend to investigate the technique in high resource languages in the future.

\section*{Conflict of interest}
The authors declare that they have no conflict of interest.

\bibliographystyle{spmpsci}
\bibliography{finetune}

\newpage

\section*{Appendix}
\label{appendix}

In this section, we provided the complete evaluation scores on the test set for all the models trained.

\begin{table}[htb]
\caption{Performance of Tag-less Back-translation compared to the baseline and standard back-translation models. BLEU Scores for each Checkpoint of the Models for Vietnamese-English NMT (best single-checkpoint and average scores are shown in \textbf{bold}).}
\label{tab:7}
\begin{tabular}{ccccccc}
\hline\noalign{\smallskip}
\multirow{2}{*}{\makecell{training step \\ (thousands)}} & \multirow{2}{*}{\emph{baseline}} & \multirow{2}{*}{\emph{standard\_bt}} & \multicolumn{2}{c}{\emph{tag-less\_bt}} & \multicolumn{2}{c}{\emph{tag-less\_bt (joint BPE)}} \\
& & & pre-train & fine-tune & pre-train & fine-tune \\
\noalign{\smallskip}\hline\noalign{\smallskip}
5 & 2.50 & 1.92 & \textbf{2.88} & & 2.87 & \\
10 & 4.25 & 3.50 & 7.54 & & \textbf{9.78} & \\
15 & 5.54 & 11.34 & 10.70 & & \textbf{13.06} & \\
20 & 6.68 & \textbf{15.83} & 12.86 & & 14.11 & \\
25 & 10.87 & \textbf{17.20} & 12.96 & & 15.03 & \\
30 & 16.89 & \textbf{19.47} & 14.07 & & 15.92 & \\
35 & \textbf{19.47} & 18.95 & 15.10 & & 15.95 & \\
40 & 20.37 & \textbf{20.81} & 15.59 & & 16.39 & \\
45 & 21.09 & \textbf{21.61} & 15.61 & & 17.08 & \\
50 & 20.13 & \textbf{22.18} & 15.99 & & 17.15 & \\
55 & 21.09 & \textbf{21.98} & 16.69 & & 17.29 & \\
60 & 21.11 & \textbf{22.35} & 16.66 & & 17.15 & \\
65 & 21.19 & \textbf{22.37} & 16.73 & & 17.37 & \\
70 & 20.17 & \textbf{23.43} & 17.63 & & 17.51 & \\
75 & 19.41 & \textbf{22.92} & 17.13 & & 17.92 & \\
80 & 20.43 & \textbf{23.41} & 17.33 & & 17.96 & \\
85 & 20.18 & \textbf{23.62} & 16.97 & & 17.94 & \\
90 & 19.80 & \textbf{23.71} & 17.85 & & 18.58 & \\
95 & 20.36 & \textbf{23.77} & 17.30 & & 18.28 & \\
100 & 19.48 & \textbf{23.73} & 17.31 & & 18.23 & \\
105 & 19.63 & \textbf{24.07} & 17.36 & & 18.60 & \\
110 & 19.77 & \textbf{23.99} & 17.34 & & 18.53 & \\
115 & - & 24.07 & 16.93 & & & \textbf{24.31} \\
120 & - & 23.98 & 17.35 & & & \textbf{24.93} \\
125 & - & 24.30 & 17.64 & & & \textbf{25.55} \\
130 & - & 24.15 & 17.62 & & & \textbf{25.79} \\
135 & - & 24.18 &  & 23.70 & & \textbf{26.21} \\
140 & - & 24.42 &  & 25.12 & & \textbf{26.11} \\
145 & - & 23.65 &  & 25.16 & & \textbf{26.33} \\
150 & - & 24.39 &  & 24.93 & & \textbf{26.23} \\
155 & - & 24.46 &  & 24.68 & & \textbf{26.21} \\
160 & - & 24.11 &  & 24.29 & & \textbf{26.45} \\
165 & - & 24.12 &  & 24.25 & & \textbf{26.53} \\
170 & - & - & - & - & & \textbf{26.40} \\
175 & - & - & - & - & & \textbf{26.17} \\
180 & - & - & - & - & & \textbf{26.23} \\
185 & - & - & - & - & & \textbf{26.21} \\
190 & - & - & - & - & & \textbf{26.16} \\
195 & - & - & - & - & & \textbf{26.21} \\ \noalign{\smallskip}\hline\noalign{\smallskip}
average & 22.22 & 25.28 & 17.96 & 25.77 & 18.59 & \textbf{26.83} \\
\noalign{\smallskip}\hline
\end{tabular}
\end{table}

\newpage

\begin{table}[htb]
\caption{Performance of Tag-less Back-translation compared to the baseline and standard back-translation models. BLEU Scores for each Checkpoint of the Models for German-English NMT (best single-checkpoint and average scores are shown in \textbf{bold}).}
\label{tab:7}
\begin{tabular}{ccccccc}
\hline\noalign{\smallskip}
\multirow{2}{*}{\makecell{training step \\ (thousands)}} & \multirow{2}{*}{\emph{baseline}} & \multirow{2}{*}{\emph{standard\_bt}} & \multicolumn{2}{c}{\emph{tag-less\_bt}} & \multicolumn{2}{c}{\emph{tag-less\_bt (joint BPE)}} \\
& & & pre-train & fine-tune & pre-train & fine-tune \\
\noalign{\smallskip}\hline\noalign{\smallskip}
5 & 2.82 & 2.13 & 0.43 & & 0.37 & \\
10 & 5.00 & 3.13 & 0.83 & & 1.11 & \\
15 & 6.30 & 4.45 & 1.09 & & 2.02 & \\
20 & 7.69 & 5.48 & 1.44 & & 2.82 & \\
25 & 9.13 & 6.30 & 1.66 & & 3.35 & \\
30 & 10.20 & 6.56 & 1.84 & & 3.80 & \\
35 & 11.23 & 7.76 & 2.12 & & 4.37 & \\
40 & 11.45 & 8.61 & 2.28 & & 4.81 & \\
45 & 11.84 & 9.76 & 2.38 & & 5.05 & \\
50 & 13.82 & 12.93 & 2.49 & & 5.14 & \\
55 & 15.73 & 16.24 & 2.77 & & 5.41 & \\
60 & 18.71 & 18.84 & 2.87 & & 5.43 & \\
65 & 19.74 & 20.21 & 2.85 & & 5.38 & \\
70 & 19.95 & 20.89 & 2.93 & &  & 21.32 \\
75 & 20.30 & 21.75 & 3.01 & &  & 23.29 \\
80 & 20.15 & 22.54 & 2.99 & &  & 24.86 \\
85 & 19.96 & 23.16 &  & 9.84 &  & 25.54 \\
90 & 19.75 & 23.44 &  & 12.74 &  & 26.32 \\
95 & 19.53 & 23.64 &  & 21.63 &  & 26.70 \\
100 & 19.51 & 24.13 &  & 23.59 &  & 26.85 \\
105 & - & 24.07 &  & 24.63 &  & 27.38 \\
110 & - & 23.94 &  & 24.84 &  & 27.67 \\
115 & - & 24.69 &  & 25.32 & & 27.82 \\
120 & - & 24.31 &  & 25.16 & & 28.06 \\
125 & - & 24.29 &  & 25.00 & & 27.91 \\
130 & - & 24.59 &  & 24.72 & & 28.01 \\
135 & - & 24.93 &  & 24.26 & & 28.03 \\
140 & - & 24.70 & - & - & & 27.98 \\
145 & - & 25.01 & - & - & & 28.28 \\
150 & - & 25.11 & - & - & & 28.26 \\
155 & - & 25.09 & - & - & & \textbf{28.31} \\
160 & - & 24.94 & - & - & & 28.24 \\
165 & - & 24.90 & - & - & & 28.15 \\
170 & - & 24.62 & - & - & & 28.22 \\
175 & - & 24.86 & - & - & - & - \\
\noalign{\smallskip}\hline\noalign{\smallskip}
average & 20.95 & 25.87 & 3.03 & 26.03 & 5.13 & \textbf{28.83} \\
\noalign{\smallskip}\hline
\end{tabular}
\end{table}

\newpage

\begin{table}[htb]
\caption{Tagged Vs Tag-less Back-translation. We used the joint BPE for implementing the tag-less approach. BLEU Scores for each Checkpoint of the Models (best single-checkpoint and average scores are shown in \textbf{bold}).}
\label{tab:8}
\begin{tabular}{ccccccc}
\hline\noalign{\smallskip}
 & \multicolumn{3}{c}{Vi-En} & \multicolumn{3}{c}{De-En} \\
\noalign{\smallskip}\cline{2-7}\noalign{\smallskip}
\multirow{2}{*}{\makecell{training step \\ (thousands)}} & \multirow{2}{*}{\emph{tagged\_bt}} & \multicolumn{2}{c}{\emph{tag-less\_bt}} & \multirow{2}{*}{\emph{tagged\_bt}} & \multicolumn{2}{c}{\emph{tag-less\_bt}} \\ \noalign{\smallskip}\cline{3-4}\cline{6-7}\noalign{\smallskip}
& & pre-train & fine-tune & & pre-train & fine-tune \\
\noalign{\smallskip}\hline\noalign{\smallskip}
5	& 1.78	& 2.87 & &	2.86	& 0.37 & \\
10	&3.64	&9.78&&	6.52&	1.11& \\
15	&9.88	&13.06&&	10.55	&2.02 &\\
20	&15.29	&14.11&&	14.74&	2.82& \\
25	&16.99	&15.03&&	16.48&	3.35& \\
30	&19.22	&15.92&&	18.12&	3.80& \\
35	&19.62	&15.95&&	19.82&	4.37& \\
40	&19.71	&16.39&&	20.49&	4.81& \\
45	&20.99	&17.08&&	21.47&	5.05& \\
50	&22.03	&17.15&&	21.88&	5.14& \\
55	&21.85	&17.29&&	22.29&	5.41& \\
60	&22.03	&17.15&&	22.75&	5.43& \\
65	&22.33	&17.37&&	23.19&	5.38& \\
70	&22.95	&17.51&&	23.67&	&21.32 \\
75	&23.05	&17.92&&	23.86&	&23.29 \\
80	&22.90	&17.96&&	24.32&	&24.86 \\
85	&23.25	&17.94&&	24.60&	&25.54 \\
90	&23.91	&18.58&&	24.35&	&26.32 \\
95	&23.87	&18.28&&	24.89&	&26.70 \\
100	&23.69	&18.23&&	25.12&	&26.85 \\
105	&24.21	&18.60&&	25.25&	&27.38 \\
110	&23.91	&18.53&&	25.41&	&27.67 \\
115	&23.90	&&24.31&	25.48&	&27.82 \\
120	&24.33	&&24.93&	25.76&	&28.06 \\
125	&24.08	&&25.55&	25.89&	&27.91 \\
130	&24.34	&&25.79&	26.04&	&28.01 \\
135	&24.17	&&26.21&	26.17&	&28.03 \\
140	&24.27	&&26.11&	26.13&	&27.98 \\
145	&24.28	&&26.33&	26.02&	&28.28 \\
150	&24.44	&&26.23&	26.56&	&28.26 \\
155	&24.15	&&26.21&	26.35&	&\textbf{28.31} \\
160	&24.50	&&26.45&	26.30&	&28.24 \\
165	&24.76	&&26.53&	26.76&	&28.15 \\
170	&24.56	&&26.40&	26.84&	&28.22 \\
175	&24.29	&&26.17&	26.65&-&-	 \\
180	&24.09	&&26.23&	26.99&-&-	\\
185	&24.34	&&26.21&	27.00&-&-	\\
190	&24.27	&&26.16&	26.90&-&-	\\
195	&24.32	&&26.21&	27.05&-&-	\\
200	&-	&-&-&	27.17&-&-	\\
205	&-	&-&-&	27.49&-&-	\\
210	&-	&-&-&	27.29&-&-	\\
215	&-	&-&-&	27.10&-&-	\\
220	&-	&-&-&	27.18&-&-	\\
225	&-	&-&-&	27.01&-&-	\\
230	&-	&-&-&	27.26&-&-	\\
235	&-	&-&-&	27.44&-&-	\\
\noalign{\smallskip}\hline\noalign{\smallskip}
avarage & 25.05 & 18.59 & 26.83 & 27.75 & 5.13 & \textbf{28.83} \\
\noalign{\smallskip}\hline
\end{tabular}
\end{table}

\newpage

\begin{table}[htb]
\caption{Pre-training on the authentic data and fine-tuning on the synthetic data for Vietnamese-English NMT. BLEU Scores for each Checkpoint of the Models}
\label{tab:9}
\begin{tabular}{ccc}
\hline\noalign{\smallskip}
\makecell{training step \\ (thousands)} & \emph{pre-train} & \emph{fine-tune} \\
\noalign{\smallskip}\hline\noalign{\smallskip}
5 & 2.50 & - \\
10 & 4.25 & - \\
15 & 5.54 & - \\
20 & 6.68 & - \\
25 & 10.87 & - \\
30 & 16.89 & - \\
35 & 19.47 & - \\
40 & 20.37 & - \\
45 & 21.09 & - \\
50 & 20.13 & - \\
55 & 21.09 & - \\
60 & 21.11 & - \\
65 & 21.19 & - \\
70 & 20.17 & - \\
75 & 19.41 & 20.00 \\
80 & - & 17.60 \\
85 & - & 17.57 \\
90 & - & 18.16 \\
95 & - & 17.62 \\
100 & - & 18.95 \\
105 & - & 17.98 \\
110 & - & 18.04 \\
115 & - & 18.63 \\
120 & - & 18.36 \\
125 & - & 18.20 \\
130 & - & 18.25 \\
135 & - & 18.46 \\
140 & - & 18.09 \\
145 & - & 18.32 \\
150 & - & 17.94 \\
155 & - & 18.21 \\
160 & - & 18.25 \\
165 & - & 18.38 \\
170 & - & 17.69 \\\noalign{\smallskip}\hline\noalign{\smallskip}
average & - & 18.91 \\
\noalign{\smallskip}\hline
\end{tabular}
\end{table}

\newpage

\begin{table}[htb]
\caption{Using different ratios of authentic to synthetic parallel data and its effect on the performance of Vietnamese-English NMT. Evaluation scores (BLEU) on the test set for each checkpoint (\emph{tag-less\_bt} colour code: BLACK -- pre-train, RED -- fine-tune)}
\label{tab:10}
\begin{tabular}{ccccccc}
\hline\noalign{\smallskip}
\multirow{2}{*}{\makecell{training step \\ (thousands)}}  & \multicolumn{3}{c}{\emph{tagged\_bt}} & \multicolumn{3}{c}{\emph{tag-less\_bt (joint BPE)}} \\
\noalign{\smallskip}
 & \textbf{1:1} & \textbf{1:3} & \textbf{1:5}  & \textbf{1:1} & \textbf{1:3} & \textbf{1:5} \\
 \noalign{\smallskip}\hline\noalign{\smallskip}
5	&2.31		&1.78		&1.87		&1.65				&2.87				&1.82 \\
10	&3.31		&3.64		&3.23		&8.68				&9.78				&6.16 \\
15	&7.35		&9.88		&4.05		&11.93			&13.06			&12.81 \\
20	&14.45	&15.29	&5.08		&13.52			&14.11			&14.20 \\
25	&17.12	&16.99	&5.71		&14.36			&15.03			&13.63 \\
30	&18.54	&19.22	&13.36	&15.24			&15.92			&16.08 \\
35	&20.45	&19.62	&16.41	&15.49			&15.95			&15.24 \\
40	&21.76	&19.71	&17.28	&15.90			&16.39			&16.34 \\
45	&21.86	&20.99	&19.20	&16.18			&17.08			&17.32 \\
50	&21.96	&22.03	&19.93	&16.05			&17.15			&16.36 \\
55	&22.87	&21.85	&20.33	&16.52			&17.29			&17.45 \\
60	&22.71	&22.03	&20.85	&16.58			&17.15			&16.99 \\
65	&23.05	&22.33	&21.62	&16.30			&17.37			&17.83 \\
70	&23.33	&22.95	&21.83	&16.82			&17.51			&16.94 \\
75	&22.84	&23.05	&21.84	&16.80			&17.92			&18.36 \\
80	&23.25	&22.90	&23.05	&16.76			&17.96			&17.86 \\
85	&23.51	&23.25	&22.56	&16.54			&17.94			&18.14 \\
90	&22.99	&23.91	&23.15	&16.47			&18.58			&18.44 \\
95	&23.06	&23.87	&23.62	&16.69			&18.28			&18.52 \\
100	&22.89	&23.69	&23.37	&\textcolor{red}{23.04}	&18.23			&18.31 \\
105	&22.96	&24.21	&23.58	&\textcolor{red}{24.05}	&18.60			&18.80 \\
110	&22.80	&23.91	&23.49	&\textcolor{red}{24.81}	&18.53			&18.42 \\
115	&22.92	&23.90	&23.90	&\textcolor{red}{24.87}	&\textcolor{red}{24.31}	&19.22 \\
120	&22.83	&24.33	&23.69	&\textcolor{red}{25.50}	&\textcolor{red}{24.93}	&19.07 \\
125	&22.29	&24.08	&23.79	&\textcolor{red}{24.86}	&\textcolor{red}{25.55}	&18.47 \\
130	&22.53	&24.34	&24.10	&\textcolor{red}{26.29}	&\textcolor{red}{25.79}	&19.44 \\
135	&22.43	&24.17	&24.01	&\textcolor{red}{25.77}	&\textcolor{red}{26.21}	&19.48 \\
140	&22.36	&24.27	&24.52	&\textcolor{red}{25.95}	&\textcolor{red}{26.11}	&19.16 \\
145	&22.34	&24.28	&24.57	&\textcolor{red}{25.94}	&\textcolor{red}{26.33}	&19.14 \\
150	&22.08	&24.44	&23.88	&\textcolor{red}{25.96}	&\textcolor{red}{26.23}	&19.01 \\
155	&22.08	&24.15	&24.62	&\textcolor{red}{25.76}	&\textcolor{red}{26.21}	&\textcolor{red}{24.82} \\
160	&21.74	&24.50	&24.49	&			-	&\textcolor{red}{26.45}	&\textcolor{red}{25.79} \\
165	&22.14	&24.76	&24.12	&			-	&\textcolor{red}{26.53}	&\textcolor{red}{26.21} \\
170	&22.73	&24.56	&25.47	&			-	&\textcolor{red}{26.40}	&\textcolor{red}{26.27} \\
175	&-		&24.29	&-		&			-	&\textcolor{red}{26.17}	&\textcolor{red}{26.51} \\
180	&-		&24.09	&-		&			-	&\textcolor{red}{26.23}	&\textcolor{red}{26.78} \\
185	&-		&24.34	&-		&			-	&\textcolor{red}{26.21}	&\textcolor{red}{26.91} \\
190	&-		&24.27	&-		&			-	&\textcolor{red}{26.16}	&\textcolor{red}{26.60} \\
195	&-		&24.32	&-		&			-	&\textcolor{red}{26.21}	&\textcolor{red}{26.57} \\
200	&-		&-		&-		&			-	&-				&\textcolor{red}{26.58} \\
205	&-		&-		&-		&			-	&-				&\textcolor{red}{26.64} \\
210	&-		&-		&-		&			-	&-				&\textcolor{red}{25.70} \\
215	&-		&-		&-		&			-	&-				&\textcolor{red}{26.29} \\
\noalign{\smallskip}\hline\noalign{\smallskip}
average & 22.73 & 25.05 & 25.47 & 26.15 & 26.83 & 27.14 \\
\noalign{\smallskip}\hline
\end{tabular}
\end{table}

\newpage

\begin{table}[htb]
\caption{Fine-tuning the tagged and standard back-translations}
\label{tab:11}
\begin{tabular}{ccc}
\hline\noalign{\smallskip}
\makecell{training step \\ (thousands)} & \emph{tagged\_bt (1:5)} & \emph{standard\_bt (1:3)} \\
\noalign{\smallskip}\hline\noalign{\smallskip}
165 & 23.38 & 24.21 \\
170 & 25.11 & 24.61 \\
175 & 25.55 & 24.87 \\
180 & 24.99 & 25.15 \\
185 & 24.79 & 24.11 \\
190 & 23.58 & 23.98 \\
195 & 23.37 & 23.36 \\
200 & 23.15 & 22.97 \\
\noalign{\smallskip}\hline\noalign{\smallskip}
average & 25.64 & 25.32 \\
\noalign{\smallskip}\hline
\end{tabular}
\end{table}

\begin{landscape}        
    \vspace*{-1.5\baselineskip}
    \captionsetup{singlelinecheck=false}
    \captionof{table}{This table shows how often a conclusion with 95\% statistical significance is made for comparing the various approaches. We used different sample sizes of 100, 500 and 1000 sentences for each of the approach on English-Vietnamese and English-German low resource NMT.}
    \label{stat-sig}
    \renewcommand{\arraystretch}{1.3}
    \begin{tabular}{lcccccccc}
	\hline\noalign{\smallskip}
	\multirow{3}{*}{System Comparison} & \multicolumn{4}{c}{Vi-En} & \multicolumn{4}{c}{De-En} \\ \cline{2-4}\cline{5-8}
	& \multirow{2}{*}{\begin{tabular}[c]{@{}c@{}}BLEU \\ difference\end{tabular}} & \multicolumn{3}{c}{Sample size} & \multirow{2}{*}{\begin{tabular}[c]{@{}c@{}}BLEU \\ difference\end{tabular}} & \multicolumn{3}{c}{Sample size} \\ \cline{3-5}\cline{7-9}
	& & 100       & 500      & 1000     & & 100       & 500       & 1000    \\ \noalign{\smallskip}\hline\noalign{\smallskip}
	Standard BT is better than baseline & 3.06 & 100\%     & 100\%    & 100\%    & 4.92 & 100\%     & 100\%     & 100\%   \\
	Tagged BT is better than baseline & 2.83 & 100\%     & 100\%    & 100\%    & 6.81 & 100\%     & 100\%     & 100\%   \\
	Tagged BT is better than Standard BT & -0.23 & 27\%      & 28\%     & 30.2\%   & 1.89 & 100\%     & 100\%     & 100\%   \\
	Tag-less BT is better than baseline & 3.55 & 100\%     & 100\%    & 100\%    & 5.09 & 100\%     & 100\%     & 100\%   \\
	Tag-less BT better than Standard BT & 0.49 & 92.8\%    & 94.6\%   & 94.8\%   & 0.16 & 87.1\%    & 87.4\%    & 84\%    \\
	Tag-less BT is better than Tagged BT & 0.72 & 98.8\%    & 98.2\%   & 98\%     & -1.72 & 0\%       & 0\%       & 0\%     \\
	\makecell[tl]{Tag-less BT (joint BPE) is better \\ \quad \quad \quad than baseline}  & 4.61 & 100\%  & 100\%    & 100\%    & 7.89 & 100\%     & 100\%     & 100\%   \\
	\makecell[tl]{Tag-less BT (joint BPE) is better \\ \quad \quad \quad than Standard BT} & 1.55 & 100\%     & 100\%    & 100\%    & 2.96 & 100\%     & 100\%     & 100\%   \\
	\makecell[tl]{Tag-less BT (joint BPE) is better \\ \quad \quad \quad than Tagged BT} & 1.78 & 100\% & 100\%   & 100\%    & 1.08 & 100\% & 100\% & 100\%  \\
	\noalign{\smallskip}\hline
     \end{tabular}
\end{landscape}

\end{document}